\pdfoutput=1

\documentclass[5p]{elsarticle} 

\usepackage{zref-xr,zref-hyperref,zref-user}
\zxrsetup{toltxlabel}
\zexternaldocument*[sup-]{supplement}
\zexternaldocument*[sup-]{supplement-z}

\usepackage{amssymb}
\usepackage{amsmath}
\usepackage{amsthm}
\usepackage{url}
\usepackage{booktabs}
\makeatletter
\def\ps@pprintTitle{%
  \let\@oddhead\@empty
  \let\@evenhead\@empty
  \def\@oddfoot{\reset@font\hfil\thepage\hfil}
  \let\@evenfoot\@oddfoot
}
\makeatother

\newtheorem{proposition}{Proposition}

\newcommand{\something}{\,\cdot\,}

\begin{document}

\emergencystretch 3em

\begin{frontmatter}

\title{Explaining Predictive Uncertainty by Exposing Second-Order Effects}

\author[tub,bifold]{Florian Bley}

\author[hhi]{Sebastian Lapuschkin}

\author[tub,bifold,hhi]{Wojciech Samek}

\author[fub,bifold]{\mbox{Grégoire Montavon}\corref{cor}}

\cortext[cor]{Corresponding author (email: \texttt{gregoire.montavon@fu-berlin.de})}

\affiliation[tub]{organization={Department of Electrical Engineering and Computer Science, Technische Universität Berlin},
            addressline={Marchstr. 23}, 
            city={Berlin},
            citysep={},
            postcode={10587}, 
            country={Germany}}

\affiliation[bifold]{
            organization={BIFOLD -- Berlin Institute for the Foundations of Learning and Data},
            city={Berlin},
            country={Germany}}

\affiliation[hhi]{
            organization={Department of Artificial Intelligence, Fraunhofer Heinrich Hertz Institute},
            addressline={Salzufer 15/16}, 
            city={Berlin},
            citysep={},
            postcode={10587}, 
            country={Germany}}

\affiliation[fub]{
            organization={Department of Mathematics and Computer Science, Freie Universität Berlin},
            addressline={Arnimallee 14}, 
            city={Berlin},
            citysep={},
            postcode={14195}, 
            country={Germany}}

\begin{abstract}
Explainable AI has brought transparency into complex ML blackboxes, enabling, in particular, to identify which features these models use for their predictions. So far, the question of explaining predictive uncertainty, i.e.\ why a model `doubts', has been scarcely studied. Our investigation reveals that predictive uncertainty is dominated by \textit{second-order effects}, involving single features or product interactions between them. We contribute a new method for explaining predictive uncertainty based on these second-order effects. Computationally, our method reduces to a simple covariance computation over a collection of first-order explanations. Our method is generally applicable, allowing for turning common attribution techniques (LRP, Gradient$\,\times\,$Input, etc.) into powerful second-order uncertainty explainers, which we call CovLRP, CovGI, etc. The accuracy of the explanations our method produces is demonstrated through systematic quantitative evaluations, and the overall usefulness of our method is demonstrated via two practical showcases.
\end{abstract}


\end{frontmatter}

\section{Introduction}

As deep learning methods make decisions in increasingly critical scenarios, measuring the degree of certainty in a prediction becomes important to avoid costly mistakes made by AI systems. In the context of autonomous driving, a model anticipating high uncertainty may choose a safer route \cite{xu2014motion} or prompt the human driver to take control in dangerous situations \cite{michelmore2020uncertainty}. When performing reinforcement learning to train a steering agent,  uncertainty estimates can allow the model to reduce speed in unfamiliar situations to avoid collisions \cite{kahn2017uncertainty}. In the context of medical diagnosis, predictive uncertainty can be used to identify out-of-distribution tissue images \cite{mehrtash2020confidence} or help to indicate in which cases consultation with an expert clinician is necessary \cite{abdar2021uncertainty}. 

\medskip

High predicted uncertainty could be due to a variety of reasons, e.g.\ the data point is different from the training distribution \cite{ovadia2019can}, or the predictive problem is locally more complex \cite{DBLP:journals/spm/MontavonBKM13}. To precisely identify and thus overcome the limitations of the model, it is vital to elucidate, e.g.\ which features of the data point, induced predictive uncertainty. Understanding a model's prediction in terms of input features, a common theme in Explainable AI \cite{Holzinger2019, samek_review_xai, klauschen_xai_pathology}, has been extensively studied for classification \cite{Bach2015, strumbelj2010efficient}, and extensions to e.g.\ regression have been proposed \cite{letzgus2021toward}. However, the question of how to explain \textit{uncertainty} of prediction models has been addressed much more scarcely.

\medskip

\begin{figure}[t]
\centering
\includegraphics[width=\linewidth]{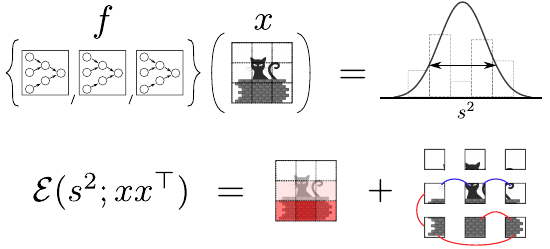}
\caption{Illustration of an ensemble model (top left), its prediction and predictive uncertainty (top right) and an illustrative cartoon example of our proposed predictive uncertainty explanation in terms of features and feature interactions (bottom). Red patches and connecting lines highlight features and feature interactions adding to predictive uncertainty; blue connecting lines highlight feature interactions decreasing predictive uncertainty.}
\label{fig:intro}
\end{figure}

In this paper, we demonstrate that explaining uncertainty, as measured by the \textit{variance} over an ensemble of predictions, differs from a standard explanation by the dominance of second-order effects. We then propose a novel second-order Explainable AI method for uncertainty predictions that accounts for these second-order effects. Our method derives from an identification of elementary product structures in the uncertainty function, based on which the second-order effects can be efficiently attributed to input features, in particular, disentangling single-feature from joint-features contributions. Our method operates on general neural network structures, including highly nonlinear ones, and integrates with existing explanation frameworks such as LRP \cite{Bach2015} and Gradient$\,\times\,$Input \cite{shrikumar2017learning}. Our method is highlighted schematically in Fig.\ \ref{fig:intro} (the way our method operates is illustrated in more detail in Fig.\ \ref{fig:overview}).
\medskip

Through systematic benchmark experiments, we demonstrate that our method yields substantially more faithful explanations of predictive uncertainty than a simple and naive application of classical Explainable AI methods to the uncertainty model. We further demonstrate via showcase examples how identifying features contributing to uncertainty via our method helps to consolidate the training data to arrive at a better, more accurate model and how our method helps to gain novel insights into a real-world dataset.

\section{Related Work}

This section reviews the literature related to our work, focusing on three main categories: the development of methods for identifying uncertain ML model decisions, approaches to explain the uncertainty behind model predictions and Explainable AI techniques that identify second- or higher-order effects in the decision behavior of general ML models.

\subsection{Estimating Uncertainty}

Predictive uncertainty can be obtained directly by modeling the output distribution (e.g.\ \cite{nix1994estimating,bishop94mixture}). However, this modeling approach applied in the context of deep neural networks often leads to overconfident uncertainty estimates \cite{guo2017calibration}. Consequently, many uncertainty estimation techniques have been proposed to remedy overconfidence in deep models and produce more accurate assessments.

In \cite{gal2016dropout}, the authors established Monte Carlo dropout (MC dropout), which applies dropout at test time to estimate uncertainty as the variance of model predictions. In \cite{teye2018bayesian}, the authors pursued a similar approach, this time, using stochastic Batch Normalization at test time. In \cite{osband2016deep}, the authors estimated uncertainty using data-dependent dropout and trained multiple prediction heads using bootstrapping. The authors in \cite{maddox2019simple} treated model parameters as Gaussian distributed and used stochastic weight averaging (SWAG) \cite{izmailov2018averaging} to estimate the approximate posterior parameters. The authors of \cite{lakshminarayanan2017simple} employed ensembles of randomly initialized deep networks to estimate predictive variation. In \cite{amini2020deep}, the authors consider a regression setting and assume, besides the conditional target distribution, a higher latent distribution, which yields the distribution parameters of the former. They then theoretically show how training the latent distribution can be regarded as an evidence-gathering process, which yields for each prediction an amount of `virtual' evidence from which the authors deduct uncertainty measures (i.e.\ a low evidence amount yields high uncertainty). For classification problems, Deep Prior Networks, as discussed in \cite{malinin2018predictive}, estimate the parameters of a latent Dirichlet distribution for the conditional class probabilities in a similar way to allow for accurate predictive uncertainty measures.

\subsection{Explaining Uncertainty}

While much research has focused on estimating the predictive uncertainty of deep models, there has been significantly less exploration into explaining predictive uncertainty, e.g.\ in terms of input features. In \cite{depeweg2017sensitivity}, the authors performed a gradient-based Sensitivity Analysis of predictive uncertainty to explain which features contribute to uncertainty. However, Sensitivity Analysis cannot account for the second-order effects in the uncertainty function. The authors in \cite{antoran2020clue} used counterfactual explanations for explaining predictive uncertainty. In particular, they trained a generative model and solved the optimization problem of finding a minimally altered sample in latent space while maximally reducing model uncertainty. While the authors could demonstrate good human interpretability of counterfactuals, their approach necessitates training a generative model. This adds additional complexity for the practitioner and renders explanations exposed to potential biases in the generator. 
The authors of \cite{watson2023explaining} theoretically explored the approach of applying Shapley Values \cite{shapley1953value, lundberg2017unified} to various uncertainty measures such as entropy and information gain and tested their approach in covariate shift and feature selection applications. While their approach is theoretically founded, it inherits some of the well-known limitations of Shapley value based explanations: Except for low-dimensional datasets or specific models, Shapley values are infeasible to compute precisely and must be approximated.

\subsection{Higher-Order Explanations}

Various methods have been proposed to extract explanations in terms of multiple interacting features. The Shapley Taylor Index \cite{shapley_taylor} extends the Shapley Value approach to highlight the contribution of feature interactions to a prediction. Likewise, Integrated Hessian \cite{integrated_hessian}
is an extension of Integrated Gradients \cite{ref_integrated_gradients} that enables an explanation in terms of feature interactions. The method operates by computing a double path integral of the predictive model's Hessian matrix towards the data point. BiLRP \cite{eberle2022building} is a second-order method that specifically addresses the explanation of product-type similarity models of pairs of data points. Predicted similarities can then be attributed robustly to pairs of input features associated to each data point. GNN-LRP \cite{schnake_xai_gnn} is a higher-order explanation method developed for Graph Neural Networks, which decomposes the model's prediction in terms of sequences of connected edges in the graph.---Our proposed approach differs from the methods above by targeting specifically the question of explaining uncertainty. It explicitly identifies the second-order structure of uncertainty predictions and allows for extracting the corresponding second-order explanations in a robust and computationally efficient manner.

\section{Proposed Method for Explaining Uncertainty}
\label{section:method}

In this section, we derive our proposed second-order approach to explain predictive uncertainty, specifically, attribute it to the input features. Our derivation leads to a simple form for the uncertainty explanation: a covariance over multiple classical (first-order) explanations, which allows us to disentangle between what is attributed onto individual and joint features.

As a starting point, we consider that the uncertainty estimate we would like to explain is given by the variance of model predictions in an ensemble of $M$ neural networks:
\begin{equation}
s^2 = \frac{1}{M}\sum_{m=1}^M \left(y_m-\bar{y}\right)^2
\label{eq:uncertainty}
\end{equation}
with $y_m$ the output of model $m$, and $\bar{y}$ the average prediction of the different models.
This formulation is general enough to include any uncertainty quantification method relying on the variance of model predictions such as deep ensembles \cite{lakshminarayanan2017simple}, MC dropout \cite{gal2016dropout}, MC batch normalization \cite{teye2018bayesian} or SWAG \cite{maddox2019simple}. We observe that the predictive variance stated in Eq.\ \eqref{eq:uncertainty} can be rewritten as a linear combination of prediction products, i.e.
\begin{equation}
s^2 = \sum_{m, m'} b_{m, m'} \cdot y_m y_{m'},
\label{eq:uncertainty-lincomb}
\end{equation}
where $b_{m,m'} = \frac1M \cdot 1_{\{m=m'\}} - \frac{1}{M^2}$ are the coefficients of the linear combination with $1_{\{\cdot\}}$ the indicator function, and where $\sum_{m, m'}$ is a nesting of two sums with each sum running over all models in the ensemble.

\medskip

We now focus on the problem of attributing the predictive variance $s^2$ to the input features. Denote by $\mathcal{E}(\cdot)$ the process of attributing what is given as argument to the input features, as defined later. The linearity observed in Eq.\ \eqref{eq:uncertainty-lincomb} lets us reduce the problem of attributing the variance $s^2$ to:
\begin{equation}
\mathcal{E}\Big(\sum_{m, m'} b_{m, m'} \cdot y_m y_{m'}\Big) = \sum_{m, m'} b_{m, m'} \cdot \mathcal{E}\big(y_m y_{m'}\big).
\label{eq:linear}
\end{equation}
In other words, the problem of attributing the predictive variance can be treated as solving simpler subproblems (the attribution of model output's products) and linearly combining the results.

\begin{figure*}[t]
    \centering
    \makebox[\linewidth][c]{
    \includegraphics[width=.875\linewidth]{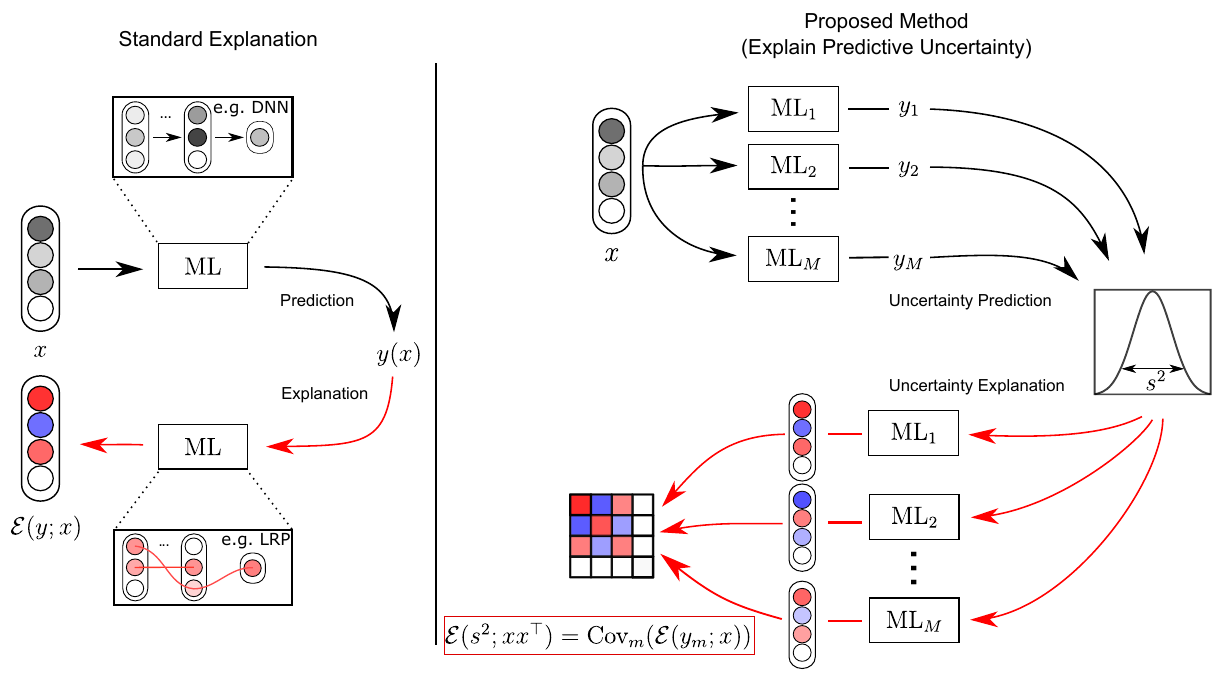}}
    \caption{Left: Classical explanation workflow, commonly used for attributing the output of a neural network model to the individual input features (elements of $x$). Right: Proposed explanation method for explaining predictive uncertainty. Predictive uncertainty (estimated by the variance over an ensemble's predictions) is attributed to elements of $xx^\top$ (a second-order explanation) by computing a covariance over explanations associated to each member of the ensemble.}
    \label{fig:overview}
\end{figure*}

Previous works, such as \cite{eberle2022building}, have demonstrated that product structures are mathematically more naturally attributed to pairs of input features. For example, the product of two linear models is a quadratic function and its monomials (products of pairs of features) form a natural basis for explanation. We denote the process of attributing to such as basis as $\mathcal{E}(\something;xx^\top)$ (compared to $\mathcal{E}(\something;x)$ for an attribution to individual features). Inspired by \cite{eberle2022building}, we propose to attribute a product of two ML outputs as the outer product of their respective first-order explanations:
\begin{equation}
\mathcal{E}\big(y_m y_{m'}\,;\,xx^\top\big) = \mathcal{E}(y_m;x) \otimes \mathcal{E}(y_{m'};x)
\label{eq:product}
\end{equation}
We provide some justification for Eq.\ \eqref{eq:product} in Section \ref{section:theoretical}, in particular, we find that it allows for maintaining useful properties of an explanation such as conservation, and zero-scores for irrelevant features. As a last step, combining \mbox{Eqs.\ \eqref{eq:linear}--\eqref{eq:product}}, we can finally observe that the overall explanation becomes the covariance over the first-order explanations of each ML model in the ensemble:
\begin{equation}\label{eq:cov}
\mathcal{E}\big(s^2\,;\,xx^\top\big) = \text{Cov}_m\big(\mathcal{E}(y_m;x)\big)
\end{equation}
The proof is given in \zref{sup-note:cov}. The covariance computation yields a matrix of size $d \times d$ encoding the contribution of each pair of features to the uncertainty. Our method is illustrated and contrasted to a classical explanation workflow in Fig.\ \ref{fig:overview}.

Note that our framework lets the user choose which first-order explanation method to use inside Eq.\ \eqref{eq:cov}. We refer to the application of our method alongside a specific first-order explanation technique by adding the prefix `Cov'. For example, if one computes $\mathcal{E}(y_m;x)$ using LRP we refer to the resulting second-order explanation of uncertainty, i.e.\ the output of Eq.\ \eqref{eq:cov}, as `CovLRP'. Likewise, if the underlying attribution technique is Gradient$\,\times\,$Input (GI), one gets `CovGI'. There is no specific restriction on the choice of underlying attribution technique. Further combinations may be considered depending on the application.

In presence of \textit{multidimensional} targets (e.g.\ output time series), predictive uncertainty can be modeled as the sum of the variances of individual output dimensions. The explanation then becomes a sum of covariances, specifically, $\mathcal{E}\big(\sum_k s_k^2\,;\,xx^\top\big) = \sum_k \mathcal{E}(s_k^2\,;\,xx^\top)$.

\section{Theoretical Properties}
\label{section:theoretical}

We show in this section that our second-order method for attributing predictive uncertainty inherits certain properties of the first-order explanation method it builds upon.

\begin{proposition}[Conservation]
If for each member $m$ of the ensemble, the corresponding output $y_m$ is attributed to the input features in a way that is conservative, i.e.\ if $\sum_i \mathcal{E}(y_m;x)_i = y_m$, then the attribution of predictive uncertainty $s^2$ according to Eq.\ \eqref{eq:cov} is also conservative, i.e.\ $\sum_{ij}{\mathcal{E}(s^2;xx^\top)_{ij} = s^2}$.
\label{proposition:conservation}
\end{proposition}
One way of proving this is to combine Eqs.\ \eqref{eq:linear} and \eqref{eq:product}, and observing that summing all elements of the resulting matrix expression yields $s^2$. The detailed proof is given in \zref{sup-note:conservation}.

\begin{proposition}[Preservation of Irrelevance]
If all models in the ensemble are invariant to a given feature $i$, and that the explanations of each prediction reflect that invariance by assigning a score of zero, i.e.\ if $\forall_{m}:\mathcal{E}(y_m;x)_i=0$, then it results that $\mathcal{E}(s^2;xx^\top)_{jk} = 0$ for all pairs $(j,k)$ where $j=i$ or $k=i$. In other words, the feature $i$ neither contributes to uncertainty on its own nor in interaction with other features.
\label{proposition:irrelevance}
\end{proposition}
This property is straightforward to demonstrate from an inspection of Eq.\ \eqref{eq:product} where features that have been attributed zero by the first-order explanations preserve their score of zero after the product operation.

\subsection{Reductions for Special Cases}

We now show that our second-order uncertainty explanations (which we can compute using Eq.\ \eqref{eq:cov}) reduce to simple and intuitive forms for special cases of models and underlying attribution techniques.

\begin{proposition}
\label{proposition:linear}
If each model in the ensemble is a linear homogeneous function of the input, i.e.\ if $y_m = w_m^\top x$ with $w_m$ the vector of parameters of model $m$ and the underlying first-order attribution method consists of the element-wise product of the weights and the input\footnote{We note that LRP, Integrated Gradients, Gradient $\times$ Input or Shapley values with a zero-valued reference point reduce to this simple attribution (cf.\ \cite{samek_review_xai}).}, i.e.\ $\mathcal{E}(y_m;x) = w_m \odot x$, then the outcome of our analysis reduces to $\mathcal{E}(s^2;xx^\top) = \mathrm{Cov}_m(w_m) \odot xx^\top$.
\end{proposition}
In other words, Proposition \ref{proposition:linear} states that, for a feature to contribute to uncertainty on its own, it must be present in the data, and the models in the ensemble must disagree on the effect of that feature for the prediction. Likewise, for two distinct features to \textit{jointly} contribute to uncertainty, both of them must be present in the data, and the models should disagree, however in some correlated manner. Specifically, some models should react \textit{positively} to the two features, and other models should disagree with the first models by reacting \textit{negatively} to the two features. Proposition \ref{proposition:linear} can be demonstrated by the chain equations: $\mathcal{E}(s^2;xx^\top) = \mathrm{Cov}_m(\mathcal{E}(y_m)) = \mathrm{Cov}_m(w_m \odot x) = \mathrm{Cov}_m(w_m) \odot xx^\top$.

We now give a closed-form expression for a classical gradient-based method, relating the proposed second-order explanation technique to the Hessian of the predicted uncertainty.

\begin{proposition}
\label{proposition:hessian}
If each model in the ensemble is a piecewise linear function of the input and if the underlying explanation method is Gradient$\,\times\,$Input, then the outcome of our analysis can be interpreted in terms of the Hessian of $s^2$ w.r.t.\ the input features, specifically, we get $\mathcal{E}(s^2;xx^\top) = {\frac12 \nabla^2 s^2(x) \odot xx^\top}$ where $\nabla^2$ denotes the Hessian operation, and $x$ is the features vector.
\end{proposition}

A proof is given in \zref{sup-note:hessian}. This result can be seen as a generalization of the form of Proposition \ref{proposition:linear}, where the Hessian operation is a local estimation of the discrepancy between the members of the ensemble.

\section{Summarizing Uncertainty Explanations}
\label{section:summarization}

In many applications of Explainable AI, the human interpreter requires as an explanation a $d$-dimensional heatmap instead of a high-dimensional matrix of feature interactions, which may be difficult to visualize and interpret. To retrieve an explanation over individual features, we consider two approaches. Let $R$ denote the original explanation computed via Eq.\ \eqref{eq:cov} and its elements given by:
$
R_{ij} = \mathcal{E}(s^2;xx^\top)_{ij}
$
A first approach to summarizing the original explanation consists of retaining only its diagonal elements, i.e.\ keeping terms of the explanation that can be unambiguously attributed to individual features, and disregarding interactions between features, i.e.\
$
    r_i^\mathrm{diag} = R_{ii}.
$
We refer to it with the suffix `diag' in our experiments. Because diagonal terms are variance computations, the resulting explanation is strictly non-negative and thus only sees features as source of uncertainty and never as inhibitors of uncertainty. An alternative summarization technique, consists of also including joint contributions and redistributing them equally to the two associated features, i.e.\
$
    r^\mathrm{marg}_i
    = R_{ii}
    + \sum_{j:j \neq i} \big(\frac12 R_{ij} + \frac12 R_{ji} \big).
$
The resulting explanation is also expressible as the column-wise sum over the original explanation of size $d \times d$. It can be further interpreted as the covariance between an input feature's relevance and the total explained output. In the experiments, we refer to this way of summarizing the explanation with the suffix `marg'.

\section{Quantitative Evaluation}

Having derived a covariance-based approach for uncertainty explanation, we will proceed with a quantitative benchmark evaluation against common (non-covariance-based) explanation techniques. We consider specifically the CovLRP and CovGI instantiations of our method, corresponding to injecting the LRP\footnote{To compute LRP explanations, we apply the generalized LRP-$\gamma$ rule \cite{montavon2019layer,Keyl2022} at each layer with $\gamma$ set heuristically to $0.2$ (cf.\ \zref{sup-note:lrp} for a definition of LRP-$\gamma$ and its generalized version).} and Gradient$\,\times\,$Input attribution techniques in Eq.\ \eqref{eq:cov}. The resulting $d\times d$ explanations are summarized using the diagonal and marginalization approaches described in Section \ref{section:summarization}. This leads overall to the following four combinations: CovLRP-diag, CovLRP-marg, CovGI-diag and CovGI-marg.

\subsection{Baselines}
We compared our proposed approach to a number of explanation methods. Our first baseline is Sensitivity Analysis (SA) \cite{depeweg2017sensitivity}. SA scores input features according to the model output's partial derivatives (for our evaluation, we apply SA to the uncertainty function of interest $s^2(x)$). A second baseline is Integrated Gradients (IG) \cite{ref_integrated_gradients}, which explains any function output by computing a path integral of the input gradients along a segment connecting the data point and some reference point. In our experiments, we use as reference the origin, which for centered data corresponds to the training data mean. We also compare our explanation to Shapley Value Sampling (SVS), which is a common sampling-based approximation of the Shapley Values in \cite{strumbelj2010efficient} and for which an implementation is readily available at Captum AI (\url{https://captum.ai/}). Lastly, we compare our approach to plain GI and LRP (i.e.\ without our proposed covariance-based formulation). Because LRP relies on a computational graph to perform attribution, we represent Eq.\ \eqref{eq:uncertainty} as an additional layer on top of the $M$ ensemble neural networks. This additional top layer consists of the activation function, i.e.\ $a(y_m) = (y_m - \bar{y})^2$ with mean $\bar{y}$ treated as constant, followed by a linear aggregation. As for CovLRP, we use the generalized LRP-$\gamma$ at each layer with $\gamma = 0.2$.

\subsection{Datasets}
\label{subsection-datasets}

We perform our evaluation on multiple regression datasets, which include the kin8nm dataset\footnote{\url{https://www.cs.toronto.edu/∼delve/data/kin/desc.html}}, as well as five datasets from the UCI Machine Learning Repository, namely, the Bias Correction\cite{uci_bias_correction}, California Housing\cite{pace1997sparse}, Wine Quality\cite{cortez2009modeling}, YearPredictionMSD\cite{uci_yearpreditionmsd} and Seoul Bike Sharing\cite{uci_Seoul_Bike} datasets. For the latter dataset, where the input data to the prediction is not strictly defined, we treated the prediction as a time series problem and used a concatenation of past and present data of a given day as input representation.

The datasets were processed in a consistent manner: they were shuffled and split into training and testing sets, with 75\% of the data allocated for training and the remaining 25\% reserved for testing\footnote{An exception is YearPredictionMSD, which comes with a predefined training-test split.}. All datasets were standardized by subtracting the training data mean and dividing by the training data standard deviation per feature.

Additionally, we considered the EPEX-FR dataset, proposed in \cite{lago2021forecasting}, which serves as a benchmark for predicting day-ahead electricity prices. This data, publicly accessible via the ENTSO-E Transparency Platform (\url{https://transparency.entsoe.eu}), spans from January 2011 to December 2016. The prediction task involves forecasting the next 24 day-ahead electricity prices in France based on the next 24 forecasted values for French electricity demand and renewable electricity production and the previous 48 hours of day-ahead prices. The data is divided temporally, with data from 2016 reserved for testing and data from 2011 to 2015 used for training, as recommended in \cite{lago2021forecasting}. As with the UCI datasets, EPEX-FR was centered and standardized.

\begin{table*}[t!]
\centering
\makebox[\linewidth]{\footnotesize\begin{tabular}{lccccccccccc}
\toprule
 & \multicolumn{2}{c}{CovLRP} & \multicolumn{2}{c}{CovGI} & LRP & \text{GI} & \text{IG} & \text{SA} & \text{SVS} \\
 & diag & marg & diag & marg & & & & & \\
\midrule
Bias Correction & \textbf{0.315} & 0.371 & \underline{0.334} & 0.512 & 0.337 & 0.485 & 0.474 & 0.738 & 0.418 \\
California Housing & \textbf{0.316} & \underline{0.328} & 0.352 & 0.394 & 0.349 & 0.398 & 0.413 & 0.523 & 0.380 \\
EPEX-FR & \textbf{0.041} & 0.046 & \underline{0.041} & 0.113 & 0.044 & 0.104 & 0.091 & 0.237 & 0.055 \\
kin8nm & 0.375 & \underline{0.372} & 0.390 & 0.404 & 0.409 & 0.410 & 0.384 & 0.482 & \textbf{0.357} \\
Seoul Bike Sharing & \textbf{0.302} & 0.322 & 0.324 & 0.363 & \underline{0.315} & 0.358 & 0.351 & 0.368 & 0.337 \\
Wine Quality & \textbf{0.396} & \underline{0.397} & 0.415 & 0.434 & 0.416 & 0.432 & 0.423 & 0.602 & 0.422 \\
YearPredictionMSD & \textbf{0.141} & 0.150 & 0.181 & 0.184 & \underline{0.146} & 0.180 & 0.193 & 0.427 & 0.151 \\
\midrule
Average & \textbf{0.269} & \underline{0.284} & 0.291 & 0.343 & 0.288 & 0.338 & 0.333 & 0.482 & 0.303 \\
\bottomrule
\end{tabular}}
\caption{AUFC scores of different explanation techniques for \textit{deep ensembles} built on a selection of datasets. Lower AUFC values indicate better, more faithful explanations. For each dataset, the best performing explanation technique is shown in bold and the second best is shown with underline.}
\label{table:pf_deep_ensemble}
\end{table*}

\subsection{Evaluation Metric}

Good explanations should be able to identify the subset of features that are most relevant to the model output. This quality of an explanation is often evaluated using pixel-flipping  \cite{Bach2015,samek-tnnls17} (which we refer in the context of our tabular data as feature-flipping). The feature-flipping procedure consists of ranking input features in order of relevance according to the explanation. One then iteratively flips (removes\footnote{To select replacement values for the removed features, we follow the approach of \cite{kauffmann2022clustering} and use kernel density estimating (KDE) inpainter with a Gaussian kernel and kernel scale optimized for maximum likelihood. The KDE inpainter resamples removed feature values based on their modeled distribution, conditioned on the non-perturbed features.}) features from most to least relevant (i.e.\ starting with the most positive scores and terminating with the most negative ones). As features are being removed one after and another, we keep track of the output of the network (in our case, the uncertainty score $s^2$), thereby creating a `flipping curve'. The faster the curve decreases, the better the explanation. We summarize this decreasing behavior using the area under the flipping curve (AUFC). In our experiment, we report the AUFC averaged over the 100 test examples with the highest predictive uncertainty.

\subsection{Results}\label{subsec:pf_results}

We begin our evaluation with a classical deep ensemble. The deep ensemble consists of 10 Multi-Layer Perceptrons (MLP) with three layers composed of 900, 600 and 300 neurons, respectively. All MLP instances are trained on the same data. We choose ReLU as an activation function in the hidden layers. All models undergo 100 training epochs. During each epoch, we evaluate the loss on a 10\% validation set held out from the training data and saved the best-performing model for final application. Uncertainty is calculated as the variance of model predictions according to Eq.\ \eqref{eq:uncertainty}. We build one such deep ensemble for each considered dataset, generate explanations with our second-order approach and our baselines, and, for each combination, perform the feature-flipping evaluation. Resulting AUFC scores for each setting are shown in Table \ref{table:pf_deep_ensemble}.

Within the proposed methods (the ones with prefix `Cov'), we observe that the diagonal summarization presented in Section \ref{section:summarization} systematically yields higher performance than the marginalization approach. This observation holds for both CovLRP and CovGI. Also, CovLRP and CovGI with the diagonal summarization both perform better than their first-order counterparts (LRP and GI). CovLRP is generally also superior to CovGI due to the higher robustness of the LRP explanation backend over GI. Our approach, specifically CovLRP with diagonal summarization, also systematically outperforms a broader range of first-order explanation techniques, in particular, SA and the computationally more expensive IG and SVS.

We then repeat the same experiment on MC dropout \cite{gal2016dropout}, a different uncertainty estimator based on generating multiple predicting instances through the \textit{dropout} mechanism. We trained one MLP of the same specifications as those used in the deep ensemble and chose a dropout rate of 0.1. We sampled 10 model variants, performing dropout at test time. Similarly to the deep ensemble, the MC dropout uncertainty is then computed as the variance of predictions using Eq.\ \eqref{eq:uncertainty}. Results are shown in Table \ref{table:pf_MC_dropout}.

\begin{table*}[t!]
\centering
\makebox[\linewidth]{\footnotesize\begin{tabular}{lccccccccccc}
\toprule
 & \multicolumn{2}{c}{CovLRP} & \multicolumn{2}{c}{CovGI} & LRP & \text{GI} & \text{IG} & \text{SA} & \text{SVS} \\
 & diag & marg & diag & marg & & & & & \\
\midrule
Bias Correction & \underline{0.301} & 0.309 & \textbf{0.301} & 0.367 & 0.344 & 0.348 & 0.345 & 0.394 & 0.360 \\
California Housing & \textbf{0.374} & \underline{0.392} & 0.396 & 0.477 & 0.433 & 0.478 & 0.435 & 0.561 & 0.489 \\
EPEX-FR & \textbf{0.061} & 0.064 & \underline{0.061} & 0.116 & 0.061 & 0.118 & 0.100 & 0.200 & 0.089 \\
kin8nm & \textbf{0.535} & \underline{0.541} & 0.579 & 0.590 & 0.632 & 0.600 & 0.577 & 0.681 & 0.603 \\
Seoul Bike Sharing & \underline{0.594} & 0.710 & \textbf{0.584} & 0.709 & 0.617 & 0.713 & 0.724 & 0.746 & 0.716 \\
Wine Quality & \textbf{0.489} & \underline{0.508} & 0.532 & 0.578 & 0.548 & 0.580 & 0.581 & 0.644 & 0.606 \\
YearPredictionMSD & \textbf{0.124} & 0.149 & \underline{0.146} & 0.200 & 0.147 & 0.184 & 0.192 & 0.344 & 0.200 \\
\midrule
Average & \textbf{0.354} & 0.382 & \underline{0.371} & 0.434 & 0.398 & 0.432 & 0.422 & 0.510 & 0.438 \\
\bottomrule
\end{tabular}}
\caption{AUFC scores of explanation techniques for \textit{MC dropout} on the considered datasets. Lower AUFC values are better. Best performance is in bold, and second best with underline.}
\label{table:pf_MC_dropout}
\end{table*}

We observe the same overall trend as in the previous experiment, with the best scores achieved by our second-order methods, in particular, the CovLRP method with diagonal summarization.

Overall, the superior performance of our proposed explanations compared to first-order methods (including computationally more expensive ones), underscores the benefit of integrating second-order effects in the explanation procedure. The higher performance of the diagonal (`diag') over the marginal (`marg') summarization may appear surprising, given that the diagonal does not include all evidence for uncertainty. One possible explanation is that feature interaction terms, which the diagonal summarization omits, are intrinsically hard to attribute to individual features; they are furthermore more sensitive to changes in feature values and thus only locally informative. As more features are being flipped, the perturbed data point moves away from the explained data point, and feature interaction information becomes unreliable. We may, therefore, see the diagonal summarization as a more global (and thereby more robust) form of explanation. 

\section{Use Case 1: Identifying Underrepresented Features in CelebA}
\label{section:usecase-celeba}

High predictive uncertainty commonly arises when a model makes predictions for data points dissimilar from the observed training data \cite{ovadia2019can}. Such covariate shift may be caused by measurement biases or insufficient and unrepresentative training data collection from the whole data population. 

As a consequence of insufficient training data collection, some input features may remain underrepresented at training time. When these features appear at test time, the model is ill-prepared to interpret their effect on the prediction task. Thus, the model prediction is unreliable and predictive uncertainty is high. In this case, explaining predictive uncertainty in terms of underrepresented features can enable the user to precisely diagnose what is missing in the current data and, subsequently, gather additional training data to improve the model. 

This section will demonstrate that our uncertainty explanation is able to reveal underrepresented high-level features at test time and how retraining on a consolidated dataset reduces uncertainty attributed to the originally underrepresented feature. To show this, we consider a setting of a model suffering from missing features in the training data. We perform fine-tuning by introducing new data points with the missing feature, improving model accuracy. We accompany this model improvement with our Explainable AI method to explain the relevant features inducing uncertainty of the original model and explain the reduction of uncertainty after fine-tuning.

In this use case, we consider the CelebA dataset\footnote{\url{https://mmlab.ie.cuhk.edu.hk/projects/CelebA.html}}, which contains over 200,000 celebrity face images and multiple annotated visual features per image. In addition, we consider the CelebA-HQ extension\footnote{\url{https://mmlab.ie.cuhk.edu.hk/projects/CelebA/CelebAMask_HQ.html}}, which adds to 30,000 CelebA images detailed segmentation masks for the visual features. We will use the 30,000 CelebA-HQ images as the test set. The CelebA dataset allows us to simulate removing a visual feature at training but not at test time, and the segmentation masks of CelebA-HQ allow us to aggregate relevance attributed to these visual features.

\begin{figure*}[t!]
\centering
    \makebox[\linewidth][c]{\includegraphics[width=.9\linewidth]{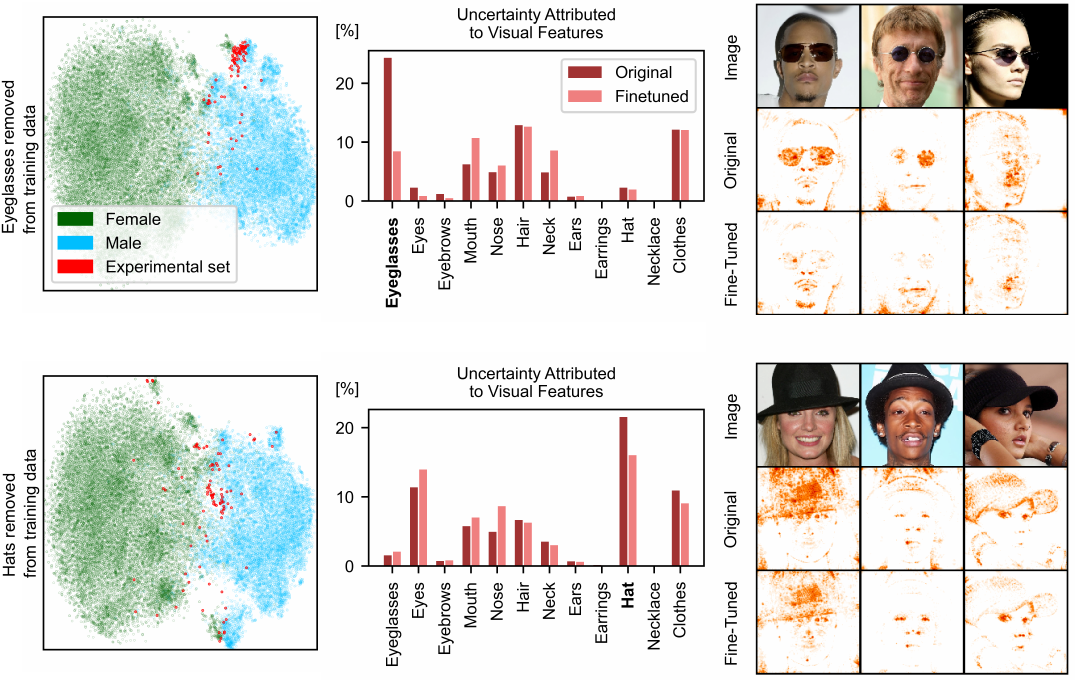}}
    \caption{Comparison of the uncertainty attribution of the original and the fine-tuned ensemble for two different underrepresented features. In the upper row, the original ensemble was trained without `eyeglass' images; in the lower row, the original ensemble was trained without `hat' images. \textit{Left:} T-SNE visualization of the original ensemble's hidden activations for the test data points, with the actual class labels colored in green-blue and the experimental set in red. \textit{Middle:} Share of uncertainty attributed to different visual features for the original and the fine-tuned ensemble, highlighting the primary role of `eyeglass' and `hat' features in reducing uncertainty. \textit{Right:} Heatmap explanations of the original and the fine-tuned ensemble for three images from the experimental data, illustrating on a pixel-wise basis the reduction of `hat' and `eyeglass' features as contributors to uncertainty.}
    \label{fig:UC1}
\end{figure*} 

We trained two ensembles of five VGG-16 \cite{simonyan2014very} networks on a male/female classification task. We trained the first ensemble on a subset of the training data, from which we removed all images exhibiting a particular feature (hats and eyeglasses). To create the second ensemble, we fine-tuned a copy of the first ensemble on the previously omitted data. Thus, while the first ensemble has no concept of the omitted features, the fine-tuned one has. Fine-tuning enhanced test accuracy from 97.9\% to 98.2\% when the `eyeglasses' feature was originally omitted, and from 98\% to 98.3\% when the hat feature was originally omitted.

Following training, we consider the subset of 100 CelebA-HQ test data points of the omitted feature with the greatest uncertainty reduction after fine-tuning. On this experimental dataset, we may expect that the ground-truth cause of the original ensemble's uncertainty lies in the underrepresented feature. A truthful uncertainty explanation is then expected to highlight this feature when explaining the uncertainty of the original ensemble. When comparing uncertainty explanations before and after fine-tuning, it is expected that fine-tuning would reduce the relevance of previously underrepresented features. To verify this, we apply CovLRP-diag\footnote{The underlying LRP explanations are computed using the generalized LRP-$\gamma$ with $\gamma = 0.1$ in the convolutional layers and $\gamma = 0.01$ in the dense layers.}, identifying for each instance the pixel-wise contributions to predictive uncertainty.


In Figure \ref{fig:UC1}, we show on the left a T-SNE embedding of the CelebA dataset, where our experimental dataset (images for which fine-tuning leads to a maximum decrease of uncertainty) is highlighted in red. On the right, we display for a selection of those input images their pixel-wise uncertainty explanations before and after fine-tuning. Pixel-wise explanations confirm that omitted features (`eyeglasses' and `hat') are a primary source of predictive uncertainty in the original ensemble, and that fine-tuning on the full data significantly reduces these source of uncertainty. Our observations are confirmed quantitatively by histograms in the middle column, measuring the uncertainty attributed to the different CelebA visual features averaged over the whole experimental data, and highlighting that the reduction in uncertainty is primarily attributable to the `eyeglasses' and `hat' features.

\section{Use Case 2: Insights into German Day-Ahead Electricity Prices}
\label{section:usecase-epex}

Practitioners of Explainable AI are often motivated by the prospect of performing data science and extracting new insights from large datasets that would otherwise be too complex for human investigation. To uncover interesting features within a dataset, an ML model can be trained on the data, and Explainable AI techniques can then be applied to highlight input features that are relevant for predicting an output \cite{slijepcevic2021explaining, Binder2021}. In the following, we demonstrate through a practically relevant use case how Explainable AI's capability to characterize input-output relations can be extended to the case where the output has the structure of an uncertainty estimator.

We considered the task of predicting German day-ahead electricity prices on the EPEX-DE dataset as in \cite{lago2021forecasting}, with a special interest for the price volatility. We organized the dataset into one target series of 24 future hourly day-ahead prices and three input channels $(x_1,x_2,x_3)$. These three input channels, representing past prices, renewable energy production, electricity demand (i.e. grid load) respectively, were organized as series of 48 entries. The past price series consists of the previous 48 hourly prices. The electricity demand and renewable production series each consists of 24 historical values and 24 forecasts for the next day.

\begin{figure*}[t!]
    \makebox[\linewidth][c]{\includegraphics[width=.85\linewidth]{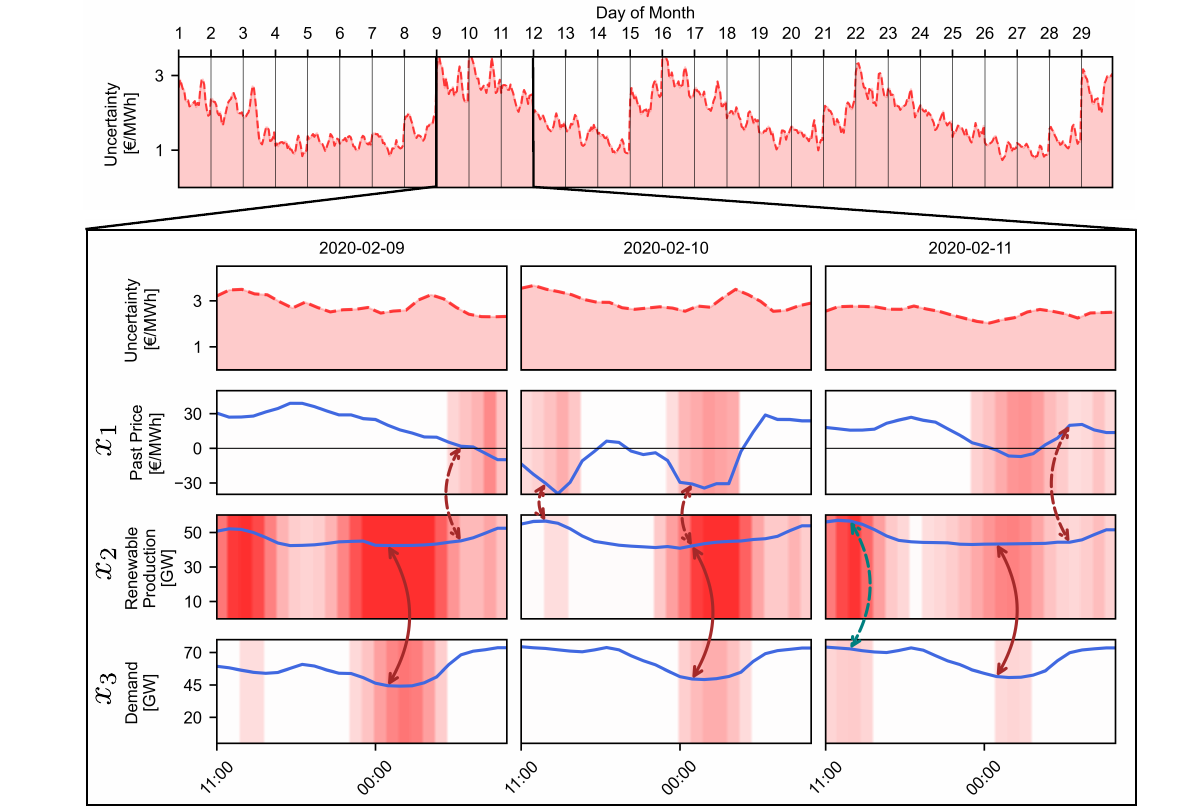}}
    \caption{Predictive uncertainty of day-ahead price prediction and uncertainty relevance analysis of three high-uncertainty days. The upper plot depicts the trained deep ensemble's hourly predictive uncertainty over the course of an entire month. The lower plot depicts for three consecutive days the predictive uncertainty and the 24 last values of the input channels $x_1,x_2,x_3$. Additionally, the CovLRP attribution of uncertainty onto these three channels is depicted in shades of red for diagonal terms, and as two-sided arrows for off-diagonal terms capturing the highest interactions. Solid connecting lines denote strong interactions and dashed connecting lines denote weaker interactions.}
    \label{fig:UC2}
    \bigskip
\end{figure*}

We applied deep ensembles to predict price volatility\footnote{While price volatility can, in principle, be derived from price time series and predicted directly via e.g.\ a neural network, such a direct approach may suffer from overconfident predictions on unseen data \cite{guo2017calibration}.}. We trained a deep ensemble of 10 convolutional neural networks with three convolutional layers, three dense layers and 24 output neurons. In all layers, we used the ReLU activation function. We used data from year 2019 for training and data from year 2020 for validation. The neural networks were randomly initialized, and we stopped training at minimal validation loss. In order to understand price volatility in terms of input features, we then applied CovLRP on the predicted uncertainty. We produced LRP explanations using the generalized LRP-$\gamma$ rule and set $\gamma = 0.3$ in the convolutional layers and $\gamma = 0.1$ in the dense layers.  Because LRP-$\gamma$ and its generalized variant always assign zero relevance to zero-valued features, potentially biasing the result of the analysis, we performed an affine transformation of the input data before training, where for each channel $x_i$, we applied the map $x_i \mapsto (1-x_i,1+x_i,2-x_i, 2+x_i)$, thereby forcing features to always have at least one non-zero value after mapping, and consequently allowing any feature value to be attributed relevance.

\begin{figure*}[t!]
  \makebox[\linewidth][c]{\includegraphics[width=0.8\linewidth]{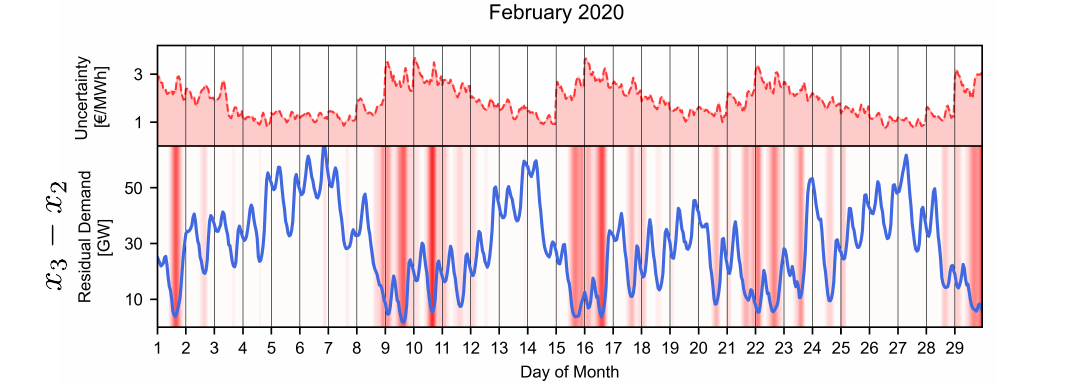}}
  \caption{Contribution of residual demand to uncertainty in February 2020. The upper plot depicts the trained deep ensemble's predictive uncertainty. The lower plot depicts residual demand, calculated as the difference between demand ($x_3$) and renewable production ($x_2$), and the shades of red show the sum of contribution to uncertainty associated to these two features. The figure shows a negative correlation between residual demand and uncertainty and the explanation's focus on low residual demand periods.}
  \label{fig_dayahead_month_aggregation}
\end{figure*}

\smallskip

As a starting point for our investigation into volatility-inducing features, we analyze the uncertainty explanation associated with three high-uncertainty days between February 9th to February 11th. During that time period, storm `Sabine' passed over Germany and caused extremely high wind power generation. Figure \ref{fig:UC2} displays the predictive uncertainty in February 2020, with a focus on the three days of storm, for which we show additionally the three input channels' time series, their respective contribution to uncertainty (CovLRP's diagonal terms), and the most significant feature interactions (CovLRP's off-diagonal terms normalized by the corresponding diagonal terms\footnote{or equivalently, off-diagonal terms of the \textit{correlation} matrix of LRP heatmaps}). The analysis of relevant feature interactions is simplified by only considering interactions of simultaneous feature values and aggregating interactions over six-hour intervals (e.g., 11h-17h, 17h-23h, etc.).

We observe the variable $x_1$ representing past prices contributes to uncertainty when those prices are negative\footnote{Negative prices are often caused by inflexible fossil energy production and low demand.}. Furthermore, while renewable production $x_2$, which is constantly high over these three days, tends to contribute strongly to uncertainty, it is especially the case during the night. Dips in expected electricity demand (variable $x_3$) during the night and weekend \footnote{February 9th, 2020 was a Sunday.}  also contribute to overall uncertainty. Moreover, strong uncertainty-inducing interactions can be observed around midnight between high renewable production ($x_2$) and low electricity demand ($x_3$).

The combined effect of high renewable production and low electricity demand on uncertainty is intriguing. It leads us to hypothesize that our method has uncovered the residual demand, defined in power markets as the difference between electricity demand ($x_3$) and renewable energy production ($x_2$), as being a main driver of uncertainty. When the expected residual demand is low and renewable sources are at peak production, fossil energy producers are compelled to reduce their output. This often results in high down-regulation costs. Depending on fuel prices, producers may choose to sell electricity below production cost to avoid these down-regulation costs. This decoupling between electricity supply and price can result in increased price volatility.

We then test our hypothesis by analyzing the relation between residual demand ($x_3-x_2$) and predictive uncertainty over an extended period of time consisting of all days of February 2020. Results are shown in Figure \ref{fig_dayahead_month_aggregation}. We observe that predictive uncertainty negatively correlates to the residual demand. Furthermore, the overall contributions of the associated features $x_2$ and $x_3$ to uncertainty is consistently high at residual demand troughs, suggesting a deeper connection between residual demand and uncertainty than a mere data correlation.---As renewable electricity production will undoubtedly increase over the next years, these results suggest that existing ML models based on demand and supply data will become insufficient to forecast day-ahead prices precisely. More generally, this analysis exemplified that the explainability of model uncertainty can help practitioners anticipate trends such as a gradual decline of predictive performance of ML models.

\section{Conclusion}
\label{section:conclusion}

In this study, we derived a novel framework for predictive uncertainty explanation in model ensembles. Our framework leverages the peculiar second-order structure of uncertainty (inherited from the variance calculation) and results in a simple yet general form consisting of computing a covariance over an ensemble's individual classical explanations. Thus, our method allows to systematically transform classical first-order explanation techniques (LRP, GI, etc.) into more powerful second-order uncertainty explainers (CovLRP, CovGI, etc.).

\medskip

In a quantitative evaluation, we demonstrated the benefits of our approach, with CovLRP achieving the highest explanation performance (as evaluated by a feature-flipping experiment), outperforming classical LRP as well as a number of other competitive baselines. We found that the superiority of CovLRP holds consistently over the multiple tabular datasets included in our benchmarks, as well as for two distinct classes of uncertainty models (deep ensembles and MC dropout). Furthermore, our approach scales linearly with the size of the ensemble and does not cause any substantial computational overhead compared to classical first-order methods.

\medskip

We then applied our framework to two practical use cases. Our first use case demonstrated that the proposed method could reveal uncertainty caused by covariate shift at test time. By identifying underrepresented features, our method can guide practitioners in collecting additional training data in a targeted fashion, ultimately improving model performance. In our second use case, we explored an electricity price dataset and focused on predictive uncertainty as a model of price volatility. Our uncertainty explanation revealed the difference between electricity demand and renewable production to be a key factor of uncertainty, also raising practical questions about the sustainability of existing predictive ML approaches in a context of gradual increase of renewable energy production.

\medskip

Our proposed uncertainty explanation method could be extended in the future towards explaining other forms of uncertainty, such as in Mixture Density Networks \cite{bishop94mixture}, where uncertainty is predicted explicitly as an output of the model. The latter may, for instance, leverage patterns in predictive uncertainty or take into account the auto-correlation of predictive uncertainty in time-series data.

\section*{Acknowledgements}

This work was funded by the German Ministry for Education and Research (refs.\ 01IS18037A, 01IS18037I, 01IS18025A); the German Research Foundation (DFG) as research unit DeSBi (KI-FOR 5363); the European Union's Horizon Europe research and innovation programme (EU Horizon Europe) as grant TEMA (101093003); the European Union's Horizon 2020 research and innovation programme (EU Horizon 2020) as grant iToBoS (965221); and the state of Berlin within the innovation support programme ProFIT (IBB) as grant BerDiBa (10174498).

\bibliographystyle{elsarticle-num}
\bibliography{literature.bib}

\begin{thebibliography}{10}
\expandafter\ifx\csname url\endcsname\relax
  \def\url#1{\texttt{#1}}\fi
\expandafter\ifx\csname urlprefix\endcsname\relax\def\urlprefix{URL }\fi
\expandafter\ifx\csname href\endcsname\relax
  \def\href#1#2{#2} \def\path#1{#1}\fi

\bibitem{xu2014motion}
W.~Xu, J.~Pan, J.~Wei, J.~M. Dolan, Motion planning under uncertainty for on-road autonomous driving, in: {ICRA}, {IEEE}, 2014, pp. 2507--2512.

\bibitem{michelmore2020uncertainty}
R.~Michelmore, M.~Wicker, L.~Laurenti, L.~Cardelli, Y.~Gal, M.~Kwiatkowska, Uncertainty quantification with statistical guarantees in end-to-end autonomous driving control, in: {ICRA}, {IEEE}, 2020, pp. 7344--7350.

\bibitem{kahn2017uncertainty}
G.~Kahn, A.~Villaflor, V.~Pong, P.~Abbeel, S.~Levine, Uncertainty-aware reinforcement learning for collision avoidance, arXiv:1702.01182 (2017).

\bibitem{mehrtash2020confidence}
A.~Mehrtash, W.~M.~W. III, C.~M. Tempany, P.~Abolmaesumi, T.~Kapur, Confidence calibration and predictive uncertainty estimation for deep medical image segmentation, {IEEE} Trans. Medical Imaging 39~(12) (2020) 3868--3878.

\bibitem{abdar2021uncertainty}
M.~Abdar, et~al., Uncertainty quantification in skin cancer classification using three-way decision-based bayesian deep learning, Computers in Biology and Medicine 135 (2021) 104418.

\bibitem{ovadia2019can}
J.~Snoek, Y.~Ovadia, E.~Fertig, B.~Lakshminarayanan, S.~Nowozin, D.~Sculley, J.~V. Dillon, J.~Ren, Z.~Nado, Can you trust your model's uncertainty? evaluating predictive uncertainty under dataset shift, in: NeurIPS, 2019, pp. 13969--13980.

\bibitem{DBLP:journals/spm/MontavonBKM13}
G.~Montavon, M.~L. Braun, T.~Krueger, K.-R. Müller, Analyzing local structure in kernel-based learning: Explanation, complexity, and reliability assessment, {IEEE} Signal Process. Mag. 30~(4) (2013) 62--74.

\bibitem{Holzinger2019}
A.~Holzinger, G.~Langs, H.~Denk, K.~Zatloukal, H.~M\"{u}ller, Causability and explainability of artificial intelligence in medicine, WIREs Data Mining and Knowledge Discovery 9~(4) (2019).

\bibitem{samek_review_xai}
W.~Samek, G.~Montavon, S.~Lapuschkin, C.~J. Anders, K.~M{\"{u}}ller, Explaining deep neural networks and beyond: {A} review of methods and applications, Proc. {IEEE} 109~(3) (2021) 247--278.

\bibitem{klauschen_xai_pathology}
F.~Klauschen, J.~Dippel, P.~Keyl, P.~Jurmeister, M.~Bockmayr, A.~Mock, O.~Buchstab, M.~Alber, L.~Ruff, G.~Montavon, K.-R. M\"{u}ller, Toward explainable artificial intelligence for precision pathology, Annual Review of Pathology: Mechanisms of Disease 19~(1) (2024).

\bibitem{Bach2015}
S.~Bach, A.~Binder, G.~Montavon, F.~Klauschen, K.-R. M\"{u}ller, W.~Samek, On pixel-wise explanations for non-linear classifier decisions by layer-wise relevance propagation, PLOS ONE 10~(7) (2015) e0130140.

\bibitem{strumbelj2010efficient}
E.~{\v{S}}trumbelj, I.~Kononenko, An efficient explanation of individual classifications using game theory, J.\ Mach.\ Learn.\ Res. 11~(1) (2010) 1--18.

\bibitem{letzgus2021toward}
S.~Letzgus, P.~Wagner, J.~Lederer, W.~Samek, K.-R. Müller, G.~Montavon, Toward explainable artificial intelligence for regression models: A methodological perspective, IEEE Signal Process. Mag. 39~(4) (2022) 40--58.

\bibitem{shrikumar2017learning}
A.~Shrikumar, P.~Greenside, A.~Kundaje, Learning important features through propagating activation differences, in: {ICML}, Vol.~70 of Proceedings of Machine Learning Research, {PMLR}, 2017, pp. 3145--3153.

\bibitem{nix1994estimating}
D.~Nix, A.~Weigend, Estimating the mean and variance of the target probability distribution, in: ICNN, Vol.~1, 1994, pp. 55--60.

\bibitem{bishop94mixture}
C.~Bishop, Mixture density networks, Tech. rep., Aston University (1994).

\bibitem{guo2017calibration}
C.~Guo, G.~Pleiss, Y.~Sun, K.~Q. Weinberger, On calibration of modern neural networks, in: {ICML}, Vol.~70 of Proceedings of Machine Learning Research, {PMLR}, 2017, pp. 1321--1330.

\bibitem{gal2016dropout}
Y.~Gal, Z.~Ghahramani, Dropout as a bayesian approximation: Representing model uncertainty in deep learning, in: {ICML}, Vol.~48 of {JMLR} Workshop and Conference Proceedings, JMLR.org, 2016, pp. 1050--1059.

\bibitem{teye2018bayesian}
M.~Teye, H.~Azizpour, K.~Smith, Bayesian uncertainty estimation for batch normalized deep networks, in: {ICML}, Vol.~80 of Proceedings of Machine Learning Research, {PMLR}, 2018, pp. 4914--4923.

\bibitem{osband2016deep}
I.~Osband, C.~Blundell, A.~Pritzel, B.~V. Roy, Deep exploration via bootstrapped {DQN} (2016) 4026--4034.

\bibitem{maddox2019simple}
W.~J. Maddox, P.~Izmailov, T.~Garipov, D.~P. Vetrov, A.~G. Wilson, A simple baseline for bayesian uncertainty in deep learning, in: NeurIPS, 2019, pp. 13132--13143.

\bibitem{izmailov2018averaging}
P.~Izmailov, D.~Podoprikhin, T.~Garipov, D.~P. Vetrov, A.~G. Wilson, Averaging weights leads to wider optima and better generalization, in: {UAI}, {AUAI} Press, 2018, pp. 876--885.

\bibitem{lakshminarayanan2017simple}
B.~Lakshminarayanan, A.~Pritzel, C.~Blundell, Simple and scalable predictive uncertainty estimation using deep ensembles, in: {NIPS}, 2017, pp. 6402--6413.

\bibitem{amini2020deep}
A.~Amini, W.~Schwarting, A.~Soleimany, D.~Rus, Deep evidential regression, in: NeurIPS, 2020.

\bibitem{malinin2018predictive}
A.~Malinin, M.~J.~F. Gales, Predictive uncertainty estimation via prior networks, in: NeurIPS, 2018, pp. 7047--7058.

\bibitem{depeweg2017sensitivity}
S.~Depeweg, J.~M. Hern{\'{a}}ndez{-}Lobato, S.~Udluft, T.~A. Runkler, Sensitivity analysis for predictive uncertainty, in: {ESANN}, 2018.

\bibitem{antoran2020clue}
J.~Antor{\'{a}}n, U.~Bhatt, T.~Adel, A.~Weller, J.~M. Hern{\'{a}}ndez{-}Lobato, Getting a {CLUE:} {A} method for explaining uncertainty estimates, in: {ICLR}, 2021.

\bibitem{watson2023explaining}
D.~S. Watson, J.~O'Hara, N.~Tax, R.~Mudd, I.~Guy, Explaining predictive uncertainty with information theoretic shapley values, arXiv:2306.05724 (2023).

\bibitem{shapley1953value}
L.~S. Shapley, A Value for n-Person Games, Princeton University Press, Princeton, 1953, pp. 307--318.

\bibitem{lundberg2017unified}
S.~M. Lundberg, S.~Lee, A unified approach to interpreting model predictions, in: {NIPS}, 2017, pp. 4765--4774.

\bibitem{shapley_taylor}
M.~Sundararajan, K.~Dhamdhere, A.~Agarwal, The shapley taylor interaction index, in: {ICML}, Vol. 119 of Proceedings of Machine Learning Research, {PMLR}, 2020, pp. 9259--9268.

\bibitem{integrated_hessian}
J.~D. Janizek, P.~Sturmfels, S.-I. Lee, Explaining explanations: Axiomatic feature interactions for deep networks, Journal of Machine Learning Research 22~(104) (2021) 1--54.

\bibitem{ref_integrated_gradients}
M.~Sundararajan, A.~Taly, Q.~Yan, Axiomatic attribution for deep networks, in: ICML, Vol.~70, 2017, pp. 3319--3328.

\bibitem{eberle2022building}
O.~Eberle, J.~B{\"{u}}ttner, F.~Kr{\"{a}}utli, K.~M{\"{u}}ller, M.~Valleriani, G.~Montavon, Building and interpreting deep similarity models, {IEEE} Trans. Pattern Anal. Mach. Intell. 44~(3) (2022) 1149--1161.

\bibitem{schnake_xai_gnn}
T.~Schnake, O.~Eberle, J.~Lederer, S.~Nakajima, K.~T. Sch{\"{u}}tt, K.~M{\"{u}}ller, G.~Montavon, Higher-order explanations of graph neural networks via relevant walks, {IEEE} Trans. Pattern Anal. Mach. Intell. 44~(11) (2022) 7581--7596.

\bibitem{montavon2019layer}
G.~Montavon, A.~Binder, S.~Lapuschkin, W.~Samek, K.-R. M{\"u}ller, Layer-Wise Relevance Propagation: An Overview, Springer, 2019, pp. 193--209.

\bibitem{Keyl2022}
P.~Keyl, M.~Bockmayr, D.~Heim, G.~Dernbach, G.~Montavon, K.-R. M\"{u}ller, F.~Klauschen, Patient-level proteomic network prediction by explainable artificial intelligence, npj Precision Oncology 6~(1) (2022) 35.

\bibitem{uci_bias_correction}
Bias correction of numerical prediction model temperature forecast, UCI Machine Learning Repository (2020).

\bibitem{pace1997sparse}
R.~K. Pace, R.~Barry, Sparse spatial autoregressions, Statistics \& Probability Letters 33~(3) (1997) 291--297.

\bibitem{cortez2009modeling}
P.~Cortez, A.~Cerdeira, F.~Almeida, T.~Matos, J.~Reis, Modeling wine preferences by data mining from physicochemical properties, Decision Support Systems 47~(4) (2009) 547--553, smart Business Networks: Concepts and Empirical Evidence.

\bibitem{uci_yearpreditionmsd}
T.~Bertin-Mahieux, {YearPredictionMSD}, UCI Machine Learning Repository (2011).

\bibitem{uci_Seoul_Bike}
{Seoul Bike Sharing Demand}, UCI Machine Learning Repository (2020).

\bibitem{lago2021forecasting}
J.~Lago, G.~Marcjasz, B.~{De Schutter}, R.~Weron, Forecasting day-ahead electricity prices: A review of state-of-the-art algorithms, best practices and an open-access benchmark, Applied Energy 293 (2021) 116983.

\bibitem{samek-tnnls17}
W.~Samek, A.~Binder, G.~Montavon, S.~Lapuschkin, K.~M{\"{u}}ller, Evaluating the visualization of what a deep neural network has learned, {IEEE} Trans. Neural Networks Learn. Syst. 28~(11) (2017) 2660--2673.

\bibitem{kauffmann2022clustering}
J.~Kauffmann, M.~Esders, L.~Ruff, G.~Montavon, W.~Samek, K.-R. Müller, From clustering to cluster explanations via neural networks, {IEEE} Trans. Neural Networks Learn. Syst. (2022) 1--15.

\bibitem{simonyan2014very}
K.~Simonyan, A.~Zisserman, Very deep convolutional networks for large-scale image recognition, in: International Conference on Learning Representations (ICLR), 2015.

\bibitem{slijepcevic2021explaining}
D.~Slijepcevic, et~al., Explaining machine learning models for clinical gait analysis, ACM Trans. Comput. Healthcare 3~(2) (2021).

\bibitem{Binder2021}
A.~Binder, et~al., Morphological and molecular breast cancer profiling through explainable machine learning, Nature Machine Intelligence 3~(4) (2021) 355–366.

\end{thebibliography}


\begin{thebibliography}{1}
\expandafter\ifx\csname url\endcsname\relax
  \def\url#1{\texttt{#1}}\fi
\expandafter\ifx\csname urlprefix\endcsname\relax\def\urlprefix{URL }\fi
\expandafter\ifx\csname href\endcsname\relax
  \def\href#1#2{#2} \def\path#1{#1}\fi

\bibitem{montavon2019layer}
G.~Montavon, A.~Binder, S.~Lapuschkin, W.~Samek, K.-R. M{\"u}ller, Layer-Wise Relevance Propagation: An Overview, Springer, 2019, pp. 193--209.

\bibitem{Keyl2022}
P.~Keyl, M.~Bockmayr, D.~Heim, G.~Dernbach, G.~Montavon, K.-R. M\"{u}ller, F.~Klauschen, Patient-level proteomic network prediction by explainable artificial intelligence, npj Precision Oncology 6~(1) (2022) 35.

\end{thebibliography}

\end{document}


\emergencystretch 3em

\begin{frontmatter}
\title{Explaining Predictive Uncertainty by Exposing Second-Order Effects\\[2mm]\large \textsc{(Supplementary Material)}}
\author{Florian Bley, Sebastian Lapuschkin, Wojciech Samek, Grégoire Montavon}
\end{frontmatter}

\noindent This supplementary material provides proofs of the main results of the paper and details of LRP rules used in our experiments.

\appendix

\section{Derivation of Eq.\ \zeqref{paper-eq:cov} of the Main Paper}
\label{note:cov}

We need to show that the expression of $\mathcal{E}(s^2)$ resulting from combining Eqs.\ \zeqref{paper-eq:linear} and \zeqref{paper-eq:product} of the main paper results in Eq.\ \zeqref{paper-eq:cov} of the main paper, in particular, one needs to show that:
\begin{align}
\sum_{m, m'} b_{m, m'} \mathcal{E}(y_m;x) \otimes \mathcal{E}(y_{m'};x) = \text{Cov}_m\big(\mathcal{E}(y_m;x)\big)
\label{eq:proof-cov-0}
\end{align}
with $b_{m,m'} = \frac1M \cdot 1_{m=m'} - \frac{1}{M^2}$. We will prove Eq.(\ref{eq:proof-cov-0}) making use of the shortcut notations $R_{im} = \mathcal{E}(y_m;x)_i$, and $\bar{R_{i}} = \frac1M \sum_m R_{im}$ with $i=1\dots d$. 
\begin{align}
    \sum_{m, m'} b_{m, m'} R_{im} R_{jm} &= \bigg(\frac1M \sum_m R_{im} R_{jm}\bigg) - \bigg(\frac1{M^2} \sum_m R_{im} \sum_{m'} R_{jm'}\bigg)\\
    &= \frac1M \sum_m \left(R_{im} R_{jm} - \bar{R}_i \bar{R}_j\right)\\
    &= \frac1M \sum_m \left(R_{im} R_{jm} -2 \bar{R}_i \bar{R}_j + \bar{R}_i \bar{R}_j \right)\\
    &= \frac{1}{M} \sum_m \left(R_{im} R_{jm} - R_{im} \bar{R}_j - R_{jm} \bar{R}_i + \bar{R}_i \bar{R}_j \right)\\
    &= \frac{1}{M} \sum_m (R_{im} - \bar{R}_i) (R_{jm} - \bar{R}_j)\\
    &=: Cov_{m} (R_{im}, R_{jm})
\end{align}

\section{Proof of Proposition \zref{paper-proposition:conservation} of the Main Paper}
\label{note:conservation}

In this section, we prove that our proposed uncertainty explanation $\mathcal{E}(s^2; xx^\top)$ satisfies the conservation property such that the sum of uncertainty relevances attributed to input feature products equals the explained predictive uncertainty $s^2$:
\begin{equation}
    \sum_{i=1}^{d} \sum_{j=1}^{d} \mathcal{E}(s^2; xx^\top)_{i, j} = s^2
\end{equation}
Proposition \zref{paper-proposition:conservation} assumes conservative first-order explanations $\mathcal{E}(y; x)$ such that we can write the following:
\begin{equation}
    \sum_{i=1}^{d} \mathcal{E}(y; x)_i = y
\end{equation}
It can be easily demonstrated that the outer product of two conservative explanations is also conservative:
\begin{align}
    \sum_{i=1}^d \sum_{j=1}^d \left(\mathcal{E}(y_m; x) \otimes \mathcal{E}(y_{m'}; x)\right)_{i, j} = & \sum_{i=1}^{d} \sum_{j=1}^{d} \mathcal{E}(y_m; x)_i \mathcal{E}(y_{m'}; x)_j\\
    = & \sum_{i=1}^d \mathcal{E}(y_m; x)_i \cdot \sum_{j=1}^{d} \mathcal{E}(y_{m'}; x)_j\\
     = & y_m \cdot y_{m'} \label{eq:outer_product_conservative}
\end{align}

By combining Eqs.\ \zeqref{paper-eq:linear} and \zeqref{paper-eq:product} of the main paper with the result of Eq.\ \eqref{eq:outer_product_conservative} we can finally prove Proposition \zref{paper-proposition:conservation}:

\begin{align}
    \sum_{i=1}^{d} \sum_{j=1}^{d} \mathcal{E}(s^2; xx^\top)_{i, j}
    &= \sum_{i=1}^{d} \sum_{j=1}^{d} \sum_{m, m'} b_{m, m'} (\mathcal{E}(y_m; x) \otimes \mathcal{E}(y_{m'}; x))_{i, j}\\
    &= \sum_{m, m'} b_{m, m'} \sum_{i=1}^d \sum_{j=1}^d (\mathcal{E}(y_m; x) \otimes \mathcal{E}(y_{m'}; x))_{i, j}\\
    &= \sum_{m, m'} b_{m, m'} y_m \cdot y_{m'}\\
    &= s^2
\end{align}

\section{Proof of Proposition \zref{paper-proposition:hessian} of the Main Paper}
\label{note:hessian}

This section proves Proposition \zref{paper-proposition:hessian} of the main paper, which states that the explanation of $s^2$ in ensembles of piece-wise linear models with CovGI involves the element-wise product of the uncertainty's Hessian, specifically:
\begin{equation}\label{eq:supp_proposition_4}
    \mathrm{Cov}_m(\mathrm{GI}(y_m)) = {\frac{1}{2} \nabla^2 s^2(x) \odot xx^\top}
\end{equation}
We will prove this by first restating the CovGI explanation of $s^2$ in terms of an element-wise product $xx^\top$ and a `CovGradient' (the covariance matrix built from the prediction's gradients). We will then show that the Hessian of uncertainty is given by twice the covariance of prediction gradients.

Starting with the CovGI explanation of $s^2$, as formulated in Eq.\ \zeqref{paper-eq:cov} of the main paper (and as it can also be expressed as a combination of Eqs.\ \zeqref{paper-eq:linear}  and \zeqref{paper-eq:product} of the main paper), we can exploit the distributive property of the Hadamard product to restate CovGI as a CovGradient multiplied by an outer product of the data:
\begin{align}
    \mathrm{Cov}_m(\mathrm{GI}(y_m))
    &= \sum_{m, m'} b_{m, m'} \text{GI}(y_m) \otimes \text{GI}(y_{m'})\\
    &= \sum_{m, m'} b_{m, m'} (\nabla y_m \odot x) \otimes (\nabla y_{m'} \odot x)\\
    &= \Big(\sum_{m, m'} b_{m, m'} (\nabla y_m \otimes \nabla y_{m'}) \Big) \odot (xx^\top)\\
    &= \text{Cov}_m (\nabla y_m) \odot xx^\top \label{eq:covgi_as_outer_product}
\end{align}
Let us now show that the Hessian of the uncertainty can also be stated in terms of this CovGradient.  We start with the expression of $s^2$ given in Eq.\ \zeqref{paper-eq:uncertainty-lincomb} of the main paper, and differentiate first with $x_i$ and then with $x_j$: 
\begin{align}
    \frac{\partial^2 s^2}{\partial x_i \partial x_j} &=
    \frac{\partial^2}{\partial x_i \partial x_j} \Big(\sum_{m, m'} b_{m, m'} y_m y_{m'}\Big)\label{eq:hessian-step1}\\
    &= \frac{\partial}{\partial x_j} \Big(\sum_{m, m'} b_{m, m'} \Big(  \frac{\partial y_m}{\partial x_i} \cdot y_{m'} + y_m \cdot \frac{\partial y_{m'}}{\partial x_i}\Big)\Big)\label{eq:hessian-step2}\\
    &= \sum_{m, m'} b_{m, m'} \Big(\frac{\partial y_m}{\partial x_i} \cdot \frac{\partial y_{m'}}{\partial x_j} + \frac{\partial y_m}{\partial x_j} \cdot \frac{\partial y_{m'}}{\partial x_i}\Big)\label{eq:hessian-step3}
\end{align}
From \eqref{eq:hessian-step1} to \eqref{eq:hessian-step2}, we have performed the first step of differentiation (with $x_i$) and used the product rule for derivatives. From \eqref{eq:hessian-step2} to \eqref{eq:hessian-step3}, we have used our assumption that the predictions $y_m$ are produced by piece-wise linear functions, and derivatives with $x_i$ can thus be seen as constants when subsequently computing the derivative with $x_j$. The Hessian matrix formed by all these second derivatives can then be compactly written as a linear combination of symmetric outer products of prediction gradients and finally as a CovGradient:
\begin{align}
    \nabla^2 s^2(x)
    &= \sum_{m, m'} b_{m, m'} (\nabla y_m \otimes \nabla y_{m'} + \nabla y_{m'} \otimes \nabla y_m) \label{eq:hessian-step4}\\
    &= 2 \cdot \sum_{m, m'} b_{m, m'} (\nabla y_m \otimes \nabla y_{m'})\label{eq:hessian-step5}\\
    &= 2 \cdot \text{Cov}_m (\nabla y_m)\label{eq:hessian-step6}
\end{align}
From \eqref{eq:hessian-step4} to \eqref{eq:hessian-step5}, we have used the fact that the two nested sums run over the same indices, allowing us to identify a symmetry and simplify the expression. From \eqref{eq:hessian-step5} to \eqref{eq:hessian-step6}, we have again identified a CovGradient structure. Finally, injecting Eq.\ \eqref{eq:hessian-step6} into Eq.\ \eqref{eq:covgi_as_outer_product} proves Eq.\ \eqref{eq:supp_proposition_4}, and thereby, proves Proposition \zref{paper-proposition:hessian} of the main paper.

\section{LRP-$\gamma$ and its Generalized Version}
\label{note:lrp}

This section describes the LRP-$\gamma$ propagation rule \cite{montavon2019layer} and a generalized version of it found in \cite{Keyl2022}. Let $j$ and $k$ be indices for neurons in two consecutive layers, with $a_j,a_k \geq 0$ their activations. The activations in these two layers are related via the equation:
$$
\textstyle a_k = \max\big(0,\sum_{0,j} a_j w_{jk} \big),
$$
where $w_{jk}$ is the weight connecting neurons $j$ and $k$ and where $\sum_{0,j}$ runs over all neurons of the first layer plus an additional neuron `$0$' with activations $a_0=1$ and a weight $w_{0k} = b_k$ representing the bias of neuron $k$.

Assume LRP has proceeded from the top of the neural network down to the layer containing neurons $k$, and has attributed to each of these neurons a relevance score $R_k$. The LRP-$\gamma$ rule propagates these scores to the layer below as:
\begin{equation}
    R_j = \sum_k \frac{a_j \cdot (w_{j, k} + \gamma \cdot w_{j, k}^{+})} {\sum_{0,j} a_j \cdot (w_{j, k} + \gamma \cdot w_{j, k}^{+})} \cdot R_k
\end{equation}
where we use the notation $(\cdot)^+ = \max(0,\cdot)$, and where $\gamma \geq 0$ is a hyperparameter to be selected. A value of $\gamma$ larger than zero emphasizes positive contributions over negative contributions in the computation of the explanation. This asymmetry in favor of positive contributions serves to enhance explanation robustness \cite{montavon2019layer}.

The simple LRP-$\gamma$ rule above assumes positive inputs and activations. A proposed generalization of LRP-$\gamma$ that can be found in \cite{Keyl2022} lifts this restriction by addressing negative inputs and activations in symmetrical ways. It assumes that neuron activations of the two consecutive layers are related as:
$$
\textstyle a_k = \rho\big(\underbrace{\big.\textstyle \sum_{0,j} a_j w_{jk}}_{z_k} \big),
$$
with $\rho$ an activation function satisfying $\rho(0)=0$. The propagation between the two consecutive layers is then defined as:
\begin{align}
 R_j&=\sum_k \frac{a_j^{+} \cdot\big(w_{jk}+\gamma w_{jk}^{+}\big)+a_j^{-} \cdot\big(w_{jk}+\gamma w_{j k}^{-}\big)}{\sum_{0, j} a_j^{+} \cdot\big(w_{jk}+\gamma w_{jk}^{+}\big)+a_j^{-} \cdot\big(w_{j k}+\gamma w_{jk}^{-}\big)} \cdot I\big(z_k>0\big) \cdot R_k \nonumber\\
& \quad +\sum_k \frac{a_j^{+} \cdot\big(w_{jk}+\gamma w_{jk}^{-}\big)+a_j^{-} \cdot\big(w_{jk}+\gamma w_{jk}^{+}\big)}{\sum_{0, j} a_j^{+} \cdot\big(w_{jk}+\gamma w_{jk}^{-}\big)+a_j^{-} \cdot\big(w_{jk}+\gamma w_{jk}^{+}\big)} \cdot I\big(z_k<0\big) \cdot R_k
\end{align}
where we use the notations $(\cdot)^+ = \max(0,\cdot)$ and  $(\cdot)^- = \min(0,\cdot)$ and where $\gamma \geq 0$ is a hyperparameter to be selected. Like for the simple LRP-$\gamma$ rule, the hyperparameter $\gamma$ is typically set to a value above zero in order to increase explanation robustness.

\bibliographystyle{elsarticle-num}
\bibliography{literature.bib}